\documentclass[journal]{IEEEtran}
\ifCLASSINFOpdf
  \usepackage[pdftex]{graphicx}
\else
\fi
\usepackage{amsmath}
\usepackage{mathtools}
\usepackage{bm}
\usepackage{adjustbox}
\usepackage{subfigure}
\usepackage{orcidlink}
\usepackage{siunitx}

\usepackage[table]{xcolor}
\usepackage{soul}
\usepackage{gensymb}
\usepackage{float}
\usepackage{authblk}
\usepackage{orcidlink}

\usepackage{algorithm}
\usepackage{algpseudocode}
\begin{document}
%
\title{Pure Inertial Navigation in Challenging Environments with Wheeled and Chassis Mounted Inertial Sensors}
\author{Dusan~Nemec~\orcidlink{0000-0003-1837-4046},
        Gal~Versano~\orcidlink{0009-0007-7766-2511},
        Itai Savin~\orcidlink{0009-0007-6911-5091},
        Vojtech~Simak~\orcidlink{0000-0001-5137-0093},
        Juraj~Kekelak~\orcidlink{0009-0008-5338-0453},
        Itzik~Klein~\orcidlink{0000-0001-7846-0654}
\thanks{D.Nemec, V. Simak and J. Kekelak are with the Department of Control and Information Systems, Faculty of Electrical Engineering and Information Technology, University of Zilina, Slovakia.}
\thanks{I. Klein, G. Versano and I. Savin are with the Hatter Department of Marine Technologies, University of Haifa, Israel.}
\thanks{}}

\maketitle

\begin{abstract}
Autonomous vehicles and wheeled robots are widely used in many applications in both indoor and outdoor settings. In practical situations with limited GNSS signals or degraded lighting conditions, the navigation solution may rely only on inertial sensors and as result drift in time due to errors in the inertial measurement. 
In this work, we propose WiCHINS, a wheeled and chassis inertial navigation system by combining wheel-mounted-inertial sensors with a chassis-mounted inertial sensor for accurate pure inertial navigation. To that end, we derive a three-stage framework, each with a dedicated extended Kalman filter. This framework utilizes the benefits of each location (wheel/body) during the estimation process. To evaluate our proposed approach, we employed a dataset with five inertial measurement units with a total recording time of 228.6 minutes. We compare our approach with four other inertial baselines and demonstrate an average position error of 11.4m, which is $2.4\%$ of the average traveled distance,  using two wheels and one body inertial measurement units. As a consequence, our proposed method enables robust navigation in challenging environments and helps bridge the pure-inertial performance gap.

\end{abstract}

%
\IEEEpeerreviewmaketitle

\section{Introduction}
\noindent 
The growing use of autonomous vehicles is largely driven by their ability to improve efficiency, increase productivity, and adapt to diverse operational needs. Advances in technology, together with the decreasing cost of electronic sensors and devices, have accelerated both research and industrial deployment of such systems \cite{borenstein1997mobile}, \cite{raj2022comprehensive}, \cite{antonyshyn2023multiple}. 
Mobile robots are employed across a broad spectrum of applications: in agriculture for crop monitoring and fruit harvesting, in logistics for indoor and outdoor delivery tasks, and in hazardous or inaccessible environments such as underground tunnels, warehouses, and search-and-rescue operations \cite{kim2024development}, \cite{baek2022ros}
Passenger cars represent one of the most widespread applications of autonomous and semi-autonomous systems  \cite{padmaja2023exploration}. Modern cars increasingly incorporate advanced driver-assistance features such as adaptive cruise control \cite{mehra2015adaptive}, lane keeping, and automated parking, all of which rely on sensor fusion and real-time decision-making. Fully autonomous cars, still under active research and development, aim to extend these capabilities to end-to-end navigation in dynamic urban environments. Cars therefore serve as both a practical domain for testing large-scale autonomy and a key driver of innovation in mobile sensing, perception, and control technologies \cite{sriram2019talk}, \cite{parakkal2017gps}. \\
\noindent 
Accurate navigation is essential for vehicles, whether they operate indoors or outdoors.  In almost every mobile robot or autonomous vehicle, inertial sensors are used to enhance performance and provide high-rate attitude and heading information. To achieve reliable localization and to mitigate the inertial drift, a variety of sensing technologies are employed, including vision-based approaches \cite{persa2001real}, \cite{servieres2021visual}, global navigation satellite systems (GNSS) \cite{li2017real}, and light detection and ranging (LiDAR) sensors \cite{cheng2018mobile}, \cite{verghese2017self}. \\ %
%
\noindent
But these extra signals are not always reliable or available. Vision methods, for example, depend heavily on the environment: in the dark they can’t see well, sudden light changes confuse them, and fast movements often blur the image, making odometry less accurate. In addition, environments with few visual features or repetitive textures hinder the ability of algorithms such as SLAM \cite{aulinas2008slam} or visual odometry \cite{nister2004visual} to function effectively, while the presence of dynamic objects introduces further challenges. GNSS has clear drawbacks. It does not work indoors or underground, and in urban canyons the signals can bounce or be blocked, often leading to poor coverage and lower accuracy. Moreover, GNSS signals can be denied, leaving the system without reliable external positioning information even in open outdoor settings. In such cases, the navigation solution must rely solely on inertial sensors, a mode referred to as pure inertial navigation, where position, velocity, and orientation are derived exclusively from the integration of accelerometer and gyroscope measurements \cite{farrell2008aided}. \\
\noindent
To cope with situations of pure inertial navigation several the literature offers three major directions:
\begin{enumerate}
    \item \textbf{Information Aiding}:  Knowledge of the vehicle's current conditions and operating environment is translated into an applicable source of information to update the navigation filter~\cite{engelsman2023information,dissanayake2002aiding, borko2018gnss}.
    \item \textbf{Periodic Trajectories}: Instead of moving in  straight line trajectories, the platform moves in a periodic motion trajectory. As a result, the inertial signal to noise ratio increases, allowing accurate positioning in short-term periods~\cite{etzion2023morpi,etzion2025snake}. 
    \item \textbf{Wheel-Mounted Sensors}: This configuration offers several benefits: it can act as a substitute for conventional wheel odometers, thereby reducing the cumulative drift typical of standard inertial navigation systems. In addition, the periodic motion of the wheel helps to attenuate constant bias errors in the sensor, leading to more stable navigation estimates~\cite{niu2121wheelImu}, \cite{wu2022wheel},\cite{wu2021comparison} .
\end{enumerate}
\noindent\\
A growing line of research investigates the use of wheel-mounted inertial sensors for mobile robots and autonomous vehicles. Studies such as \cite{niu2121wheelImu}, \cite{collin2014mems}, \cite{mezentsev2019design} have proposed dead-reckoning approaches that exploit the placement of inertial sensors at the hub of a non-steering wheel. Also a work that used SLAM by using wheel-mounted inertial sensors \cite{wu2022wheel}. Further improvement can be achieved by integrating the wheel-mounted IMU with GNSS \cite{wu2025gins}, but it requires GNSS signal to be available. \\
\noindent
\noindent To provide a reliable and accurate navigation solution for mobile robots and cars in challenging environmentalists, where cameras or GNSS signals are unavailable, we propose WiCHINS a wheeled and chassis inertial navigation system. To this end, we introduce a pure inertial three-stage framework. In each stage, a dedicated extended Kalman filter (EKF) is applied. Our approach leverages inertial measurements from two wheel-mounted sensors combined with a chassis-mounted inertial measurement unit (IMU). Our framework's three stages are:
\begin{enumerate}
    \item \textbf{WheelEKF:} Wheel speed and angle are estimated with an EKF for each wheel using wheel-mounted gyroscope readings for the prediction step and wheel-mounted accelerometer for the update step. 
    \item \textbf{OriEKF:} The body orientation is estimated from the chasses-mounted IMU. To this end, an EKF is designed with the gyroscope readings propagating the system and the accelerometer measurements are used to update the filter. 
    \item \textbf{PosEKF:} The wheel-mounted accelerometers, fused across wheels are used to obtain the forward acceleration. Then, by using double integration the body velocity and position are estimated. 
\end{enumerate}
This division allows us to better control which sensor affects which state variable within the entire process, aimed in providing accurate pure inertial positioning.\\
\noindent We evaluated our method on a real-world data collected from a passenger car. The experiments were conducted with a Skoda Roomster equipped with wheel-mounted IMU sensors, while a GNSS-RTK system served as the ground truth (GT). In total, 11 trajectories were recorded during the field experiments, amounting to 228 minutes of data across all IMUs (25.4 minutes per one IMU). We show an average 2D position error of 3.26 m using two wheel and one body IMUs over four other inertial baselines. As a consequence, the method enables robust navigation in challenging environments and helps bridge the pure-inertial performance gap. \\
\noindent The rest of this paper is organized as follows: Section~\ref{sec:PA} gives the Wheel-Mounted inertial navigation, and odometry equations. Section~\ref{sec:PA} presents our proposed three stage framework. Next, in Section~\ref{sec:AR}, a detailed description of the dataset, and results is provided. Finally, Section~\ref{sec:con} derives the conclusions of this research.
\section{Mobile Robot Inertial Navigation and Odometry}\label{sec:PM}
In this section we briefly describe the wheel-mounted inertial equations as well as wheel-based odometry. 
\subsection{Wheel-Mounted Inertial Navigation}
\begin{figure}[h]
    \centering
    \includegraphics[width=0.8\linewidth]{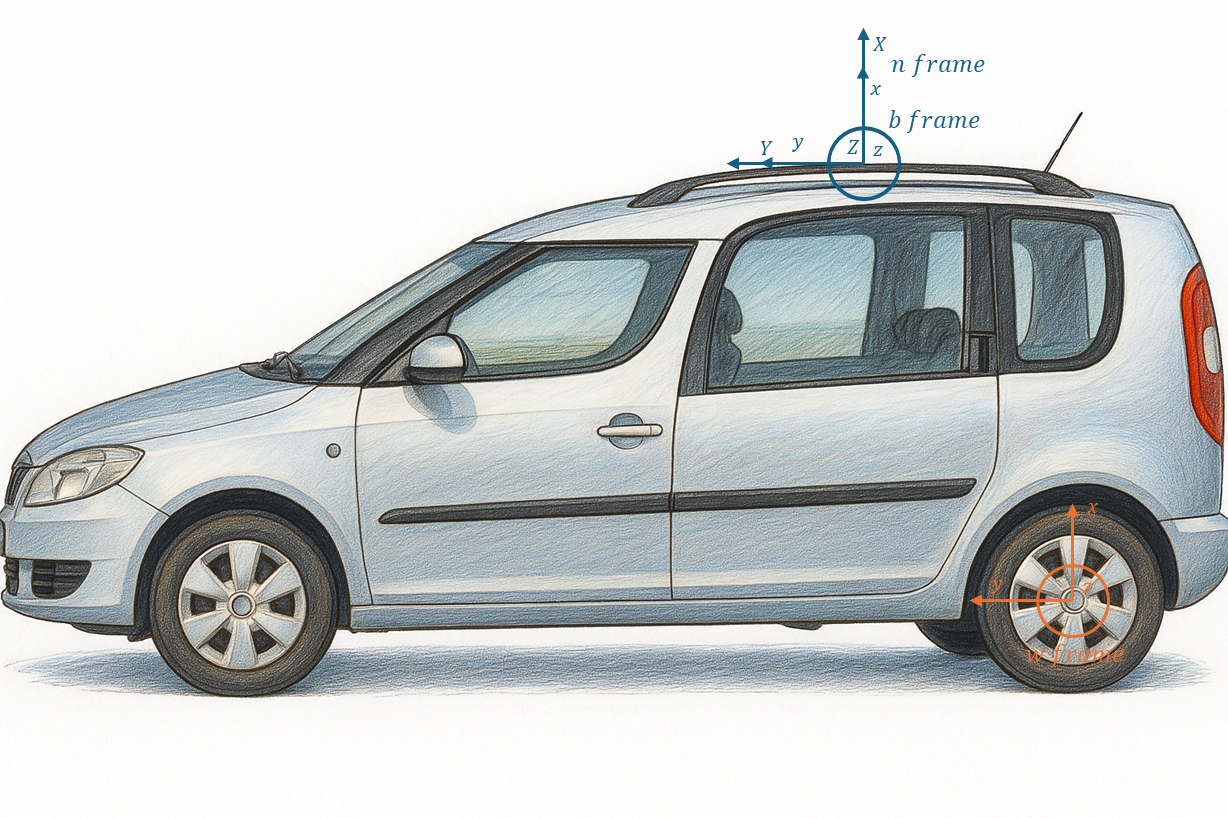}
    \caption{An illustration of a Skoda Roomster with reference frames: the wheel has the w frame, the car roof has the body frame (lowercase), and the car roof also shows the navigation frame (uppercase).}
    \label{fig:skoda_frames}
\end{figure}

\noindent Inertial navigation system (INS) dynamics are commonly expressed in the navigation frame, using north–east–down (NED) coordinates \cite{groves2015principles}. Within this article, we use 3 types of coordinates:
\begin{itemize}
    \item \textbf{Navigation frame} - static world frame of reference, its origin is identical to the starting point of the robot, oriented according to the NED convention (N = $x$-axis goes north, E = $y$-axis goes east, D = $z$-axis goes down). All variables measured with respect to the world frame will be denoted with upper index $^n$.
    \item \textbf{Body frame} - frame bound with the body of the robot, with $x$-axis going forward, $y$-axis going right and $z$-axis going down. All variables measured with respect to the body frame will be denoted with upper index $^b$.
    \item \textbf{Wheel frame} - bound with the wheel-mounted IMU, with $x$-axis going radial, $y$-axis going tangential and $z$-axis going axial (see \ref{fig:skoda_frames}). All variables measured with respect to the $i$-th wheel frame will be denoted with upper index $^{wi}$.
\end{itemize}
\noindent The time derivative of the position vector can be written as:
\begin{equation} 
\dot{\bm{p}}^{n} = \bm{v}^{n}
\label{eq:pos_eq_1}
\end{equation}

\noindent where $\bm{p}^n$ denotes the position vector in the navigation frame ($n$-frame), and $\bm{v}^n$ represents the velocity vector in the same frame. The time derivative of the velocity is then given by:
\begin{equation}
\dot{\bm{v}}^{n} = \bm{T}^{n}_{b}\,\bm{f}^{b} + \bm{g}^{n}
\label{eq:pos_eq_2}
\end{equation}

\noindent where $\bm{g}^n$ is the gravity vector in the $n$-frame, while $\bm{T}_b^n$ is the transformation matrix that converts data from the body frame to the $n$-frame. The term $\bm{f}^b$ denotes the specific force vector as measured in the body frame.\\The transformation matrix rate of change is given by:
\begin{equation}
\dot{\bm{T}}_n^b = \bm{T}_n^b \boldsymbol{\Omega}^{b}
\label{eq:tran_eq}
\end{equation}

\noindent where $\boldsymbol{\Omega}^{b}$ denotes the skew-symmetric matrix constructed from the angular rates provided by the gyroscope, expressed in the body frame. Since our study relies on low-cost inertial sensors the influences of the Earth’s rotation and transport rate are disregarded in (\ref{eq:pos_eq_2}) and (\ref{eq:tran_eq}).\\
For wheeled robot and vehicle navigation, we make use of a local body-fixed frame ($b$-frame), defined at the vehicle's starting position, with its axes oriented along the north–east–down directions.

\noindent In the wheel-mounted configuration, computing the inertial navigation solution requires transforming the raw measurements from the wheel-mounted IMU into the body frame. To place this transformation in context, three coordinate frames are considered. The navigation frame (\(n\)) is a static reference frame whose origin coincides with the robot’s starting position and is oriented according to the NED convention (north, east, down). The body frame (\(b\)) is fixed to the robot, with its x-axis pointing forward, y-axis pointing to the right, and z-axis pointing downward. Finally, the wheel frame (\(w_i\)) is attached to the IMU on the \(i\)-th wheel, defined with the x-axis radial, y-axis tangential, and z-axis axial. Figure~\ref{fig:skoda_frames} illustrates the relationship among these frames and highlights the transformation between the wheel frame, the body frame and the world frame.

\noindent
The transformation involves calculating the phase angle of the wheel at each timestamp. As defined in \cite{collin2014mems}, the phase angle is:
\begin{equation}
\alpha(t)= \int_{t}\omega_z(\tau) d\tau
\label{eq:tran2}
\end{equation}
where \(\omega_z\) is the measured angular velocity and $t$ is the time. \\
\noindent 
By using (\ref{eq:tran2}) we calculate the transformation between the $w$-frame to the $b$-frame:
\begin{equation}
\bm{T}_{b}^{w} = \begin{bmatrix} 
\cos\alpha(t) & \sin\alpha(t) & 0 \\
-\sin\alpha(t) & \cos\alpha(t) & 0 \\
0 & 0 & 1 
\end{bmatrix}
\label{eq:tran1}
\end{equation}
\noindent 
Thus the angular velocity vector in the $b$-frame is
\begin{equation}
\bm{\omega}^b= \bm{T}_{w}^{b} \bm{\omega}^w
\label{eq:ang_vel}
\end{equation}
\noindent 
where \(\boldsymbol{\omega^w}\) is the measured angular velocity of the inertial sensor mounted on the wheel. \\
\noindent In the same manner, the specific force vector expressed in the b-frame is:
\begin{equation} \label{eq:insWheelAccel}
\bm{f}^b= \bm{C}_{w}^{b} \bm{f}^w
\end{equation}
\noindent where \(\bm{f}^w\) is specific force measured from the inertial sensor mounted on the wheel.
To compute the navigation solution, the angular velocity vector (\ref{eq:ang_vel}) and  the specific force vector (\ref{eq:insWheelAccel}) are substituted into (\ref{eq:pos_eq_1})-(\ref{eq:tran_eq}).
\subsection{Wheel-based Odometry}
\noindent Odometry is a well-known technique for localizing wheeled mobile robots and vehicles. It calculates the velocity (and angular velocity) of the vehicle's body from measured angular velocities of individual wheels. It assumes no-skid conditions on all measured wheels:
\begin{equation} \label{eq:odometryCond}
    \bm{v}^b+\bm{\omega}^b \times \bm{r}^b_i-\Omega_iR_i\bm{n}^b_i=\bm{0}
\end{equation}
\noindent where $\bm{r}^b_i$ is the (constant) position of the $i$-th wheel hub (centre) in the body frame, $R_i$ is the effective rolling radius of $i$-th wheel and $\bm{n}^b_i$ is the unit direction vector of the $i$-th wheel:
\begin{equation} \label{eq:wheelDirVector}
    \bm{n}^b_i = \begin{bmatrix}\cos{\beta_i} & \sin{\beta_i} & 0\end{bmatrix}^T
\end{equation}
\noindent By rewriting the condition (\ref{eq:odometryCond}) into a matrix  form we can obtain:
\begin{equation} \label{eq:odometryProblem}
    \begin{bmatrix}
        \Omega_i R_i \cos{\beta_i} \\ \Omega_i R_i \sin{\beta_i}
        \\ \vdots
    \end{bmatrix} = \underbrace{\begin{bmatrix}
        1 & 0 & -y^b_i \\
        0 & 1 & x^b_i \\
        \vdots & \vdots & \vdots
    \end{bmatrix}}_{\bm{D}} \cdot \begin{bmatrix}
        v^b_x \\ v^b_y \\ \omega^b_z
    \end{bmatrix}
\end{equation}
\noindent The (partial) velocity vector can be calculated by a pseudo-inversion:
\begin{equation} \label{eq:forwardKinematics}
    \begin{bmatrix}
        v^b_x \\ v^b_y \\ \omega^b_z
    \end{bmatrix} = \left(\bm{D}^T\bm{D}\right)^{-1}\bm{D}^T \begin{bmatrix}
        \Omega_i R_i \cos{\beta_i} \\ \Omega_i R_i \sin{\beta_i}
        \\ \vdots
    \end{bmatrix}
\end{equation}
The eq. (\ref{eq:forwardKinematics}) represent the forward kinematics solution for any wheeled vehicle with standard wheels under no-skid conditions. After transformation of the velocity vector into navigation frame we may estimate the position of the vehicle by simple discrete (e.g. forward Euler) integrator. \\
The inverse kinematics solution computes the wheel speeds and steering angles from known velocity vector:
\begin{equation} \label{eq:invKinematicsSteering}
    \beta_i = \text{atan2}\left(v^b_y + \omega^b_z x^b_i,v^b_x - \omega^b_z y^b_i\right)
\end{equation}
\begin{equation} \label{eq:invKinematicsSpeed}
    \Omega_i = \frac{\left(v^b_x - \omega^b_z y^b_i\right)\cos{\beta_i}+ \left(v^b_y + \omega^b_z x^b_i\right) \sin{\beta_i}}{R_i}
\end{equation}
\section{Proposed approach}\label{sec:PA}
\begin{figure*}
    \centering
    \includegraphics[width=1.0\linewidth]{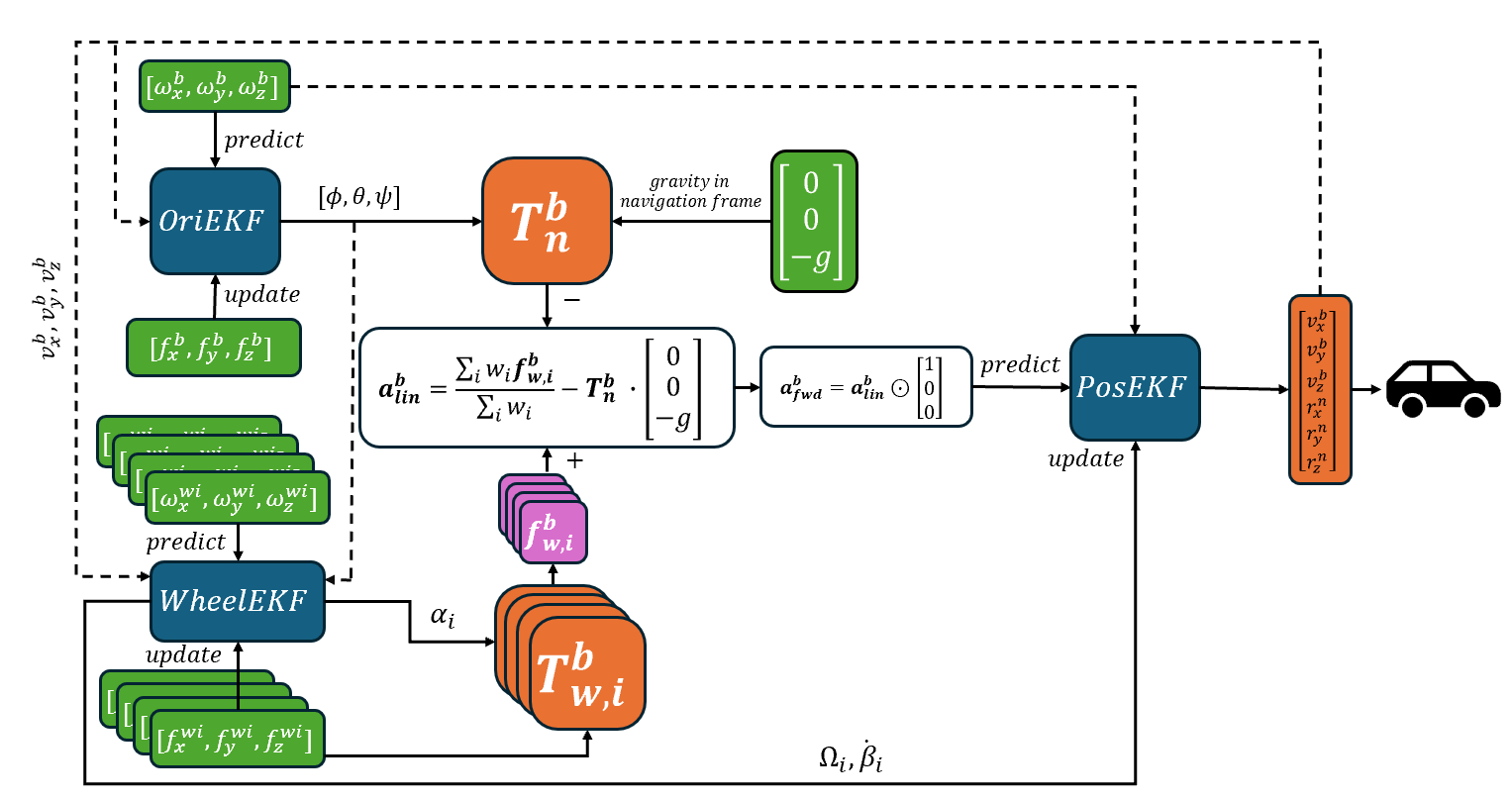}

    \caption{Block diagram our WiCHINS framework. Green blocks represent sources of information (sensors or constants), orange blocks are operations, blue blocks are EKF-based estimators, with state input measurements denoted as "predict", and output measurements as "update". Dashed lines represent signals passed to EKF as parameters.}
    \label{fig:main-scheme}
\end{figure*}

\noindent
To provide a reliable and accurate navigation solution for mobile robots and cars in scenarios where cameras or GNSS signals are unavailable, we introduce a navigation algorithm based on wheel-mounted IMUs combined with a chassis-mounted IMU. All inertial readings are fused in novel three stage approach to estimate the vehicle’s position. Our proposed approach three stages are:
\begin{enumerate}
    \item \textbf{WheelEKF:} Wheel speed and angle are estimated with an EKF for each wheel using wheel-mounted gyroscope readings for prediction step and wheel-mounted accelerometer for update step. The main idea is to use multi-wheel odometry based on wheel-mounted gyroscopes as the source of absolute information about the robot’s velocity vector, while the wheel-mounted accelerometers provide information about the velocity changes. Such setup has the potential to detect and compensate odometry errors (mainly wheel skidding, sliding and wheel diameter changes). Since the accelerometers are rotating with the wheels, their offset cancels within each turn.
    \item \textbf{OriEKF:} The body orientation (namely the roll, pitch, and yaw angles) is estimated from the body-mounted IMU. To this end, an EKF is designed with the gyroscope readings propagating the system, and accelerometer measurements are used to update the filter. The result of this stage is the world to body transformation matrix.
    \item \textbf{PosEKF:} The wheel-mounted accelerometers, fused across wheels, and subtracted gravity in the body frame (using the output of the OriEKF), are used to obtain forward acceleration. Then, by using double integration, the body velocity and position are estimated. For the update step, we use the wheel-mounted gyroscopes. To this end, inverse kinematics is used to compute the wheel speed (rotation around wheel's main axis) from the apriori estimate of the body's velocity, and then the body’s angular velocity is rotated into the wheel frame and added to it. This estimated wheel angular velocity assumes no-skid conditions. It is compared with the measurements from wheel-mounted gyroscope and used by EKF to correct the apriori estimate of the body's velocity.
\end{enumerate}
This division allows us to better control which sensor affects each state variable within the entire process aimed in providing accurate pure inertial positioning. Our complete three stage approach is shown in Fig \ref{fig:main-scheme}.

\noindent In general, one of the main challenges is the estimation of the wheels’ rotation angles, which are required for the transformation of wheel accelerometer readings from the wheel frame into the body frame. Measured specific force reflects both wheel rotation angle (through gravitational acceleration), and the body velocity changes. All sensors are affected by the body attitude changes. Our experiments showed that single EKF using all sensors at once tends to diverge in the estimation of the body attitude, which subsequently severely degrades the quality of the estimated body velocity vector. Hence, we decided to split the estimation task into the three stages as described above. In the following, we elaborate on each stage of our proposed approach. 
\subsection{OriEKF -  Body Orientation estimation}
\noindent 
The body orientation state is expressed in terms of three Euler angles referred to as roll $\phi$, pitch $\theta$, and yaw $\psi$. In our implementation the state vector is:
\begin{equation}\label{eq:oriEKF_State}
    \bm{x}_{\text{Ori}} = \begin{bmatrix} \phi & \theta & \psi \end{bmatrix}^{T}
\end{equation}
Since wheeled robots and ground vehicles in general have limited pitch, the gimbal lock condition (90 degrees upward or downward) cannot occur under normal conditions, and the estimated state vector may safely contain Euler angles directly, without the risk of numerical instability. The sensor fusion of gyroscope, accelerometer, and optionally magnetometer, based on some variant of extended Kalman filter, is a widely applied approach for the orientation estimation see e. g. \cite{deMarina2012ukf},\cite{tong2018marg}\cite{liang2023qukf}. \\
\noindent
In our setup, we have used a body-mounted IMU unit consisting of 3-axis gyroscope, and 3-axis accelerometer mounted parallel to the body frame axes. The wheel-mounted IMU sensors are not participating in this stage. 
The EKF prediction phase is based on propagating the Euler angle rates, using on the measured (and calibrated) gyro readings: 
\begin{equation} \label{eq:rpyPredict}
\begin{bmatrix} 
    \hat{\phi} \\ 
    \hat{\theta} \\ 
    \hat{\psi}
\end{bmatrix} = \begin{bmatrix} 
    \phi \\ 
    \theta \\ 
    \psi
\end{bmatrix} + \begin{bmatrix*}[c]
    1 & \dfrac{s_xs_y}{c_y} & \dfrac{c_xs_y}{c_y} \\
    0 & c_x & -s_x \\
    0 & \dfrac{s_x}{c_y} & \dfrac{c_x}{c_y}
\end{bmatrix*} \cdot \bm{\omega}^b_{cal} \Delta{t}
\end{equation}
where $\Delta{t}$ is the sampling period of the gyroscope, $\bm{\omega}^b_{cal}$ is the calibrated gyroscope readings, and we use the following notation: 
\begin{align} \label {eq:subs1}
    s_x = \sin{\phi} && c_x = \cos{\phi} \\
    s_y = \sin{\theta} && c_y = \cos{\theta} \\
    s_z = \sin{\psi} && c_z = \cos{\psi}
\end{align} 
Body orientation correction (EKF update phase) is made using body-mounted accelerometer and magnetometer calibrated readings with the combined measurement, $\bm{z}_{\text{Ori}}$, is defined as:
\begin{equation} \label{eq:OriMeas}
\bm{z}_{\text{Ori}} = \begin{bmatrix} \bm{f}^b_{cal} & \bm{b}^b_{cal}\end{bmatrix}^T 
\end{equation}
where $\bm{f}^b_{cal}$ is the measured accelerometer readings and 
$\bm{v}^b_{cal}$ is the measured magnetometer readings.
In our setup, however, the magnetometer readings were not available, hence only specific force measurement was used in \eqref{eq:OriMeas}.
\begin{equation} \label{eq:rpyUpdate}
    \hat{\bm{f}}^b = \hat{\bm{T}}^b_n \cdot \begin{bmatrix}0 & 0 & -g\end{bmatrix}^T + \bm{\omega}^b \times \bm{v}^b
\end{equation}
where: $\hat{\bm{f}}^b$ is the specific force measurement from the 3-axial body accelerometer, $g$ is the local gravity vector, $\bm{\omega}$ is the measured angular velocity vector of the body, $\bm{v}^b$ is the estimated linear velocity vector of the robot rotated into body frame, and $\hat{\bm{T}}^b_n$ is an apriori estimate of the rotational matrix $\bm{T}^b_n$ transforming from navigation-to-body coordinates:
\begin{equation} \label{eq:world2body_matrix}
    \bm{T}^b_n = \begin{bmatrix}
        c_y c_z & c_y s_z & -s_y \\
        s_x s_y c_z - c_x s_z & s_x s_y s_z + c_x c_z & s_x c_y \\
        c_x s_y c_z + s_x s_z & c_x s_y s_z - s_x c_z & c_x c_y        
    \end{bmatrix}
\end{equation}
%
\subsection{WheelEKF - Wheel State Estimation}
\noindent 
The "wheelEKF" block in Fig. \ref{fig:main-scheme} contains a set of EKFs, one for each wheel, with the following state vector for wheel $i$:
\begin{equation}\label{eq:wheelEKFstate}
    \bm{x}_{\text{Wheel,i}} = \begin{bmatrix} \Omega_i & \alpha_i & \dot{\beta}_i & \beta_i \end{bmatrix}^{T}
\end{equation}
where $\Omega_i$ is wheel rotation speed, $\alpha_i$ is rotation angle, $\dot{\beta}_i$ is the steering angular velocity, and $ \beta_i$ is steering angle for steerable wheels. In our implementation, all four wheels state vectors  were vectorized into a single EKF, with the drawback of using larger dimensions for filter matrices. \\
\noindent
The EKF prediction phase receives the calibrated wheel-mounted gyroscope readings $\bm{\omega}_{cal} = \begin{bmatrix}\omega^{wi}_x & \omega^{wi}_y & \omega^{wi}_z\end{bmatrix}$ and uses the following set of the rate-of-change equations of the state:  
\begin{equation} \label{eq:wheelPredictWheelSpeed}
    \hat{\Omega}_i = \sigma_i \omega^{wi}_z
\end{equation}
\begin{equation} \label{eq:wheelPredictWheelAngle}
    \hat{\alpha}_i = \alpha_i + \hat{\Omega}_i \Delta{t}
\end{equation}
\begin{equation} \label{eq:wheelPredictSteerSpeed}
    \hat{\dot{\beta}}_i = q_i (\omega^{wi}_x\sin\hat{\alpha}_i + \sigma_i \omega^{wi}_y \cos\hat{\alpha}_i - \omega^b_z)
\end{equation}
\begin{equation} \label{eq:wheelPredictSteerAngle}
    \hat{\beta}_i = q_i \text{angle}(\beta_i + \hat{\dot{\beta}}_i \Delta{t})
\end{equation}
where $\hat{\Omega}_i$ is the predicted rotational speed of the $i$-th wheel, $\sigma_i$ is the sign of the wheel (+1 for wheels on left side, -1 for wheels on the right side of the robot), $q_i$ is equal to 1 when $i$-th wheel is steerable otherwise 0, $\hat{\alpha}_i$ is the predicted rotational angle of $i$-th wheel, $\hat{\beta}_i$, and $\hat{\dot{\beta}}_i$ are the predicted steering angle and steering angular velocity of $i$-th wheel, respectively, and
\begin{equation} \label{eq:angleFcn}
    \text{angle}(x) = x - 2\pi\lfloor\frac{x + \pi}{2\pi}\rfloor
\end{equation}
removes the period from angle in radians, effectively mapping it into $(-\pi, \pi)$ interval. \\ 
\noindent
Wheel state correction is provided by wheel-mounted accelerometer measurement:
\begin{equation}
    \bm{z}_{\text{Wheel,i}}=\bm{f}^{w}_{cal}
\end{equation}
where 
\begin{equation} \label{eq:wheelUpdate}
    \hat{\bm{f}}^{w} = \hat{\bm{T}}^{wi}_b \cdot (\bm{f}^b + \bm{\omega}^b \times (\bm{\omega}^b \times \bm{r}^b_i)) - \begin{bmatrix}
        \rho_i \Omega_i^2 & 0 & 0
    \end{bmatrix}^T
\end{equation}
$\bm{f}^b$ is the a posteriori estimate of the specific force in the body frame (consisting of body linear acceleration, and gravity computed from corrected Euler angles using eq. (\ref{eq:rpyUpdate}), $\rho_i$ is the distance between the wheel-mounted IMU and the wheel hub (zero when mounted in the middle of the wheel), and $\hat{\bm{T}}^{wi}_b$ is an apriori estimate of the rotation matrix $\bm{r}^b_i$, which defines the transformation from the body frame to $i$-th wheel frame \cite{farrell2008aided}:
\begin{equation} \label{eq:body2wheel_matrix}
    \begin{adjustbox}{max width=0.85\columnwidth}
        $\bm{T}^{wi}_b = \begin{bmatrix}
            \cos{\beta_i} \cos{\alpha_i} & \sin{\beta_i} \cos{\alpha_i} & \sin{\alpha_i} \\
            -\sigma_i \cos{\beta_i} \sin{\alpha_i} & -\sigma_i \sin{\beta_i} \sin{\alpha_i} & \sigma_i \cos{\alpha_i} \\
            \sigma_i \sin{\beta_i} & -\sigma_i \cos{\beta_i} & 0
        \end{bmatrix}$
    \end{adjustbox}
\end{equation}
The term $\bm{\omega}^b \times (\bm{\omega}^b \times \bm{r}^b_i)$ in (\ref{eq:wheelUpdate}) represents the centrifugal acceleration observed at the wheel hub caused by rotation of the robot's body. The term $\rho_i \Omega_i^2$ represents the centrifugal acceleration caused by wheel's own rotation. We neglect the tangential acceleration components (caused by angular acceleration of body and wheel), because they can be seen as a short-term noise. The orientation of the platform is entering the block "wheelEKF" as an external parameter, hence the wheel-mounted accelerometers does not influence the estimate of robot's attitude and angular velocity.
\subsection{PosEKF - position and velocity estimation}
\noindent 
The PosEKF provides the platform's position and velocity. To this end, the state vector is:
\begin{equation}
    \bm{x}_{\text{Pos}} = \begin{bmatrix} v_x^b & v_y^b & v_z^b & r_x^n & r_y^n & r_z^n \end{bmatrix}^{T}
\end{equation}
where $v_j$ are the body linear velocity components along the body axes and $r_j$ are the position components expressed in the local navigation frame.\\
\noindent
For the prediction of the body velocity, the wheel-mounted accelerometers are used. 
First, the centrifugal acceleration $\rho_i (\omega^{wi}_z)^2$ caused by wheel's own rotation is compensated,  then the specific force vector need to be converted to body frame (block "body to frame" in Fig. \ref{fig:main-scheme}). After that, the centrifugal acceleration $\bm{\omega}^b \times (\bm{\omega}^b \times \bm{r}^b_i)$ as a result of the body rotation is subtracted:
\begin{equation}\label{eq:fbi}
    \bm{f}^b_{w,i} = \bm{T}^b_{w,i} \cdot \left(\bm{f}^{w}_{cal} + \begin{bmatrix}\rho_i (\omega^{wi}_z)^2 & 0 & 0\end{bmatrix}^T\right) - \bm{\omega}^b \times (\bm{\omega}^b \times \bm{r}^b_i)
\end{equation}
where $\bm{f}^{w}_{cal}$ is the calibrated specific force vector. 
Since a wheeled platform has usually two or more wheels, these compensated measurements need to be fused together. The most straight-forward solution for homogeneous sensor fusion is a weighted average. The weights of each wheel can be adjusted according to the sensor noise RMS, $a_{i(RMS)}$, using the rule \cite{nemec2023homo}:
\begin{equation} \label{eq:weightRule}
    w_i = \frac{1}{a_{i(RMS)}^2}
\end{equation}
Note that if we assume all wheel-mounted sensors have the same noise RMS, the homogeneous sensor fusion becomes a simple arithmetical average.
After the fusion, the gravity vector (rotated to the body frame by the block "world to body" in Fig. \ref{fig:main-scheme}) can be subtracted. Since the wheeled vehicle usually accelerates only forward (skid acceleration is negligible and the radial centrifugal acceleration has been already compensated), we may take use only the forward component of the acceleration ($x$-axis in body frame) in \eqref{eq:fbi}:
\begin{equation} \label{eq:linAccelBody}
    \bm{a}^b_{lin} = \frac{\sum_i{w_i \bm{f}^b_{w,i}}}{\sum_i{w_i}} - \bm{T}^b_n \cdot \begin{bmatrix}0 & 0 & -g\end{bmatrix}^T
\end{equation}
with $w_i$ defined in \eqref{eq:weightRule}.
The forward acceleration vector, expressed in the body frame, is then:
\begin{equation} \label{eq:fwdAccelBody}
    \bm{a}^b_{fwd} = \bm{a}^b_{lin} \odot \begin{bmatrix}1 & 0 & 0\end{bmatrix}^T
\end{equation}
where $\odot$ denotes element-wise vector multiplication. The linear acceleration vector is then used 
for prediction of the linear velocity in the body frame:
\begin{equation} \label{eq:posePredictVel}
    \hat{\bm{v}}^b = \bm{v}^b + \bm{a}^b_{fwd} \Delta{t}
\end{equation}
The linear velocity is corrected using the wheel-mounted gyroscopes:
\begin{equation}
    \bm{z}_{\text{Pos}} = \begin{bmatrix}\bm{\omega}^{w}_{cal,i}\end{bmatrix}
\end{equation}
The wheel speeds are predicted from a-priori estimate of the linear velocity $\hat{\bm{v}}^b$, and the angular velocity $\bm{\omega}^b$ using inverse kinematics of the wheeled robot (\ref{eq:invKinematicsSpeed}):
\begin{equation} \label{eq:poseUpdateWheelSpeed}
    \begin{bmatrix}\hat{\Omega}_i\end{bmatrix} = \text{InverseKinematics}(\hat{\bm{v}}^b,\bm{\omega}^b)
\end{equation}
Then, the body angular velocity and steering rate is rotated into wheel frame, and added to the wheel speed:
\begin{equation} \label{eq:poseUpdateGyro}
    \hat{\bm{\omega}}^{wi} = \begin{bmatrix}0 & 0 & \sigma_i \hat{\Omega}_i\end{bmatrix}^T + \bm{T}^{wi}_b \cdot \left( \bm{\omega}^b + \begin{bmatrix}0 & 0 & \dot{\beta}_i\end{bmatrix}^T\right)
\end{equation}
The measurement phase of the PosEKF implements both (\ref{eq:poseUpdateWheelSpeed} and \ref{eq:poseUpdateGyro}). Once the velocity estimate is corrected by poseEKF, the new position of the robot is predicted by:
\begin{equation} \label{eq:posePredictPos}
    \bm{r}^n \gets \bm{r}^n + \bm{T}^n_b \cdot \bm{v}^b \Delta{t}
\end{equation}
%
\section{Analysis and Results}\label{sec:AR}
\subsection{Dataset}\label{sec:Data}
\begin{figure}[h]
    \centering
    \includegraphics[width=0.8\linewidth]{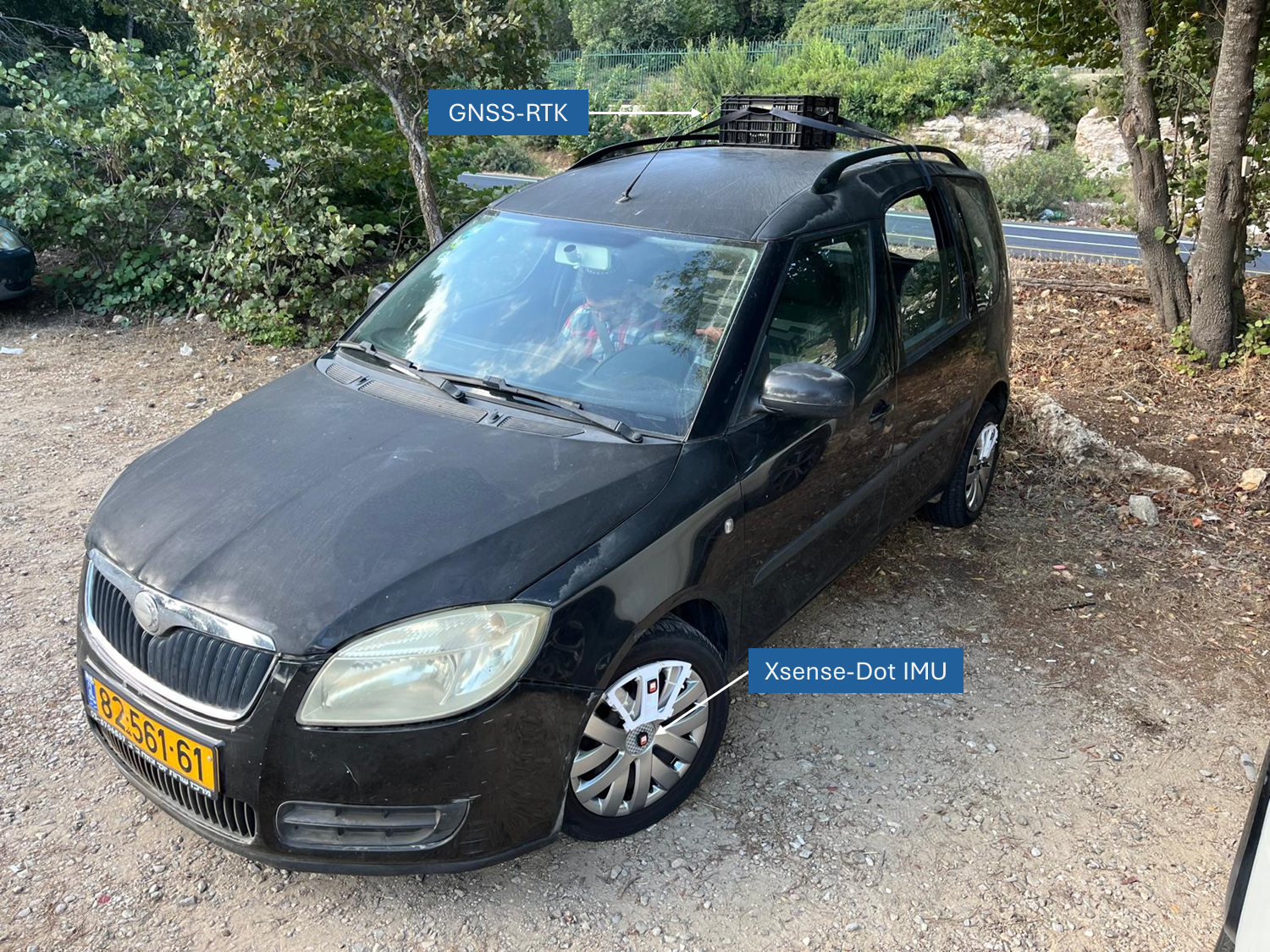}
    \caption{Skoda Roomster equipped with wheel-mounted IMUs and a GNSS-RTK on the roof~\cite{ourdata26}}.
    \label{fig:skoda_room}
\end{figure}

\noindent 
To evaluate our proposed approach we employ our recent real-world recorded dataset~\cite{ourdata26}. It includes recordings made with a Skoda Roomster car equipped with Movella Xsens DOT IMUs \cite{movella_dot}, one mounted on each wheel, was used to record our dataset. The DOT software enables synchronization between the IMUs. The associated noise and bias values of the accelerometers and gyroscopes are presented in Table~\ref{tab:gyro_accel}. Both sensors provide measurements at a rate of 120~Hz. In addition, the car was equipped with an MRU-P \cite{inertial_labs} with RTK-GNSS capabilities to provide the ground truth (GT) position at a sampling rate of 5~Hz. The car with the mounted sensors and supporting electronics is shown in Figure~\ref{fig:skoda_room}. The wheel diameter is set to fixed value $R_i = \qty{0.295}{m}$. \\
\begin{table}[h!]
\centering
\small 
\caption{Xsense DOT IMU specifications \cite{movella_dot}.}
\begin{tabular}{|c|c|c|}
\hline
    & \textbf{Gyro} & \textbf{Accelerometer} \\ 
\hline
    \textbf{Bias}   & $\qty{10}{\deg \cdot h^{-1}}$   & $\qty{0.03}{mg}$         \\ 
\hline
    \textbf{Noise}   & $\qty{0.007}{\deg \cdot s^{-1}}/\sqrt{\text{Hz}}$ & $\qty{120}{\mu g /\sqrt{\text{Hz}}}$ \\ \hline
\end{tabular}
\label{tab:gyro_accel}
\end{table}
\noindent 
During field experiments, 11 trajectories were recorded, with a total duration of 42 minutes for a single IMU and 383 minutes for the five IMUs (four mounted on the wheels and one on the car). Each recording contains raw inertial measurements along with the corresponding GT position. Table \ref{tab:trialDurations} provides the length, duration, and maximum velocity of each trajectory.
\begin{table}[h!]
\centering
\small 
\caption{Dataset trajectories duration, distance, and maximum velocity \cite{ourdata26}.}
\label{tab:trialDurations}
\begin{tabular}{|c|c|c|c|c|c|c|}
\hline
\textbf{Trajectory} & \textbf{Duration [s]} & \textbf{Distance [m]} & \textbf{Max Vel. [m/s]} \\ \hline
1 & 161.24 & 549.09 & 7.43\\ \hline
2 & 128.58 & 527.99 & 7.78\\ \hline
3 & 137.8 & 553.57 & 10.39\\ \hline
4 & 139.79 & 555.89	& 10.35\\ \hline
5 & 212.33 & 936.11	& 14.25\\ \hline
6 & 88.21 & 207.28 & 22.30\\ \hline
7 & 152.76 & 544.96	& 20.44\\ \hline
8 & 121.68 & 392.92 & 6.14\\ \hline
9 & 105.26 & 311.06	& 5.09\\ \hline
10 & 112.87 & 190.84 & 3.98\\ \hline
11 & 163.81 & 452.78 & 4.36\\ \hline
\textbf{Total} & \textbf{1524.34} & \textbf{5222.48} & -\\ \hline
\textbf{Average} & \textbf{138.58} & \textbf{474.77} & \textbf{10.23}\\ \hline
\end{tabular}
\end{table}

\subsection{Performance Metrics}
\noindent To evaluate our results, we use three metrics as described below. 
\begin{enumerate}
    \item \textbf{Position root mean square error (PRMSE)}: The PRMSE of the 3D position compares the estimated position of the vehicle in the navigation frame with the GNSS-RTK ground truth (GT):
    \begin{equation} \label{eq:rmse}
    \text{PRMSE} = \sqrt{\frac{1}{N} \sum_{k=1}^{N} ||\textbf{p}^n_k-\hat{\textbf{p}}^{n}_k||^{2}}
    \end{equation}
    where $\hat{\textbf{p}}^n_k$ are the 3D estimated positions, and $\textbf{p}^n_k$ is the GT positions obtained from the GNSS measurements. 
    \item \textbf{Velocity root mean square error (VRMSE)}: The VRMSE of the 3D velocity compares the estimated velocity of the vehicle in the navigation frame with the GNSS-RTK GT:
    \begin{equation} \label{eq:vrmse}
    \text{VRMSE} = \sqrt{\frac{1}{N} \sum_{k=1}^{N} ||\textbf{v}^n_k-\hat{\textbf{v}}^{n}_k||^{2}}
    \end{equation}
    where $\hat{\textbf{v}}^n_k$ are the 3D estimated velocities, and $\textbf{v}^n_k$ is the GT velocities obtained from the GNSS-RTK measurements. 
    \item \textbf{Total distance error (TDE)}: The TDE metric quantifies the the position error relative to the trajectory length:  
    \begin{equation}
    \text{TDE} (\%) = \frac{\text{{PRMSE}}}{\text{{L}}} \times 100
    \label{tde}
    \end{equation}
    where $L$ is the length of the trajectory.
\end{enumerate}
%

\subsection{Comparison with Existing Inertial Methods} \label{sec:Comp}
\noindent 
We evaluate two implementations of our approach and compare them with three other inertial baselines as described below.
\begin{enumerate}
    \item \textbf{2WiCHINS}: Our proposed method, as described in Section~\ref{sec:PA}, using two wheel-mounted and one body-mounted IMU. 
    \item \textbf{4WiCHINS}: Our proposed method, as described in Section~\ref{sec:PA},  using four wheel-mounted and one body-mounted IMU.
    \item \textbf{ODO}: The standard odometry using only two wheel-mounted gyroscopes (instead of encoders) as described in \eqref{eq:odometryProblem} - \eqref{eq:forwardKinematics}.
    \item \textbf{WMI}: Wheel-mounted inertial navigation using single wheel IMU as described in \eqref{eq:pos_eq_1} - \eqref{eq:insWheelAccel}.
    \item \textbf{CMI}: Chassis-mounted inertial navigation using the chassis-mounted IMU as described in \eqref{eq:pos_eq_1} - \eqref{eq:pos_eq_2}.
\end{enumerate}
\subsection{Results}
\begin{figure*}
    \centering
    \includegraphics[width=0.48\linewidth]{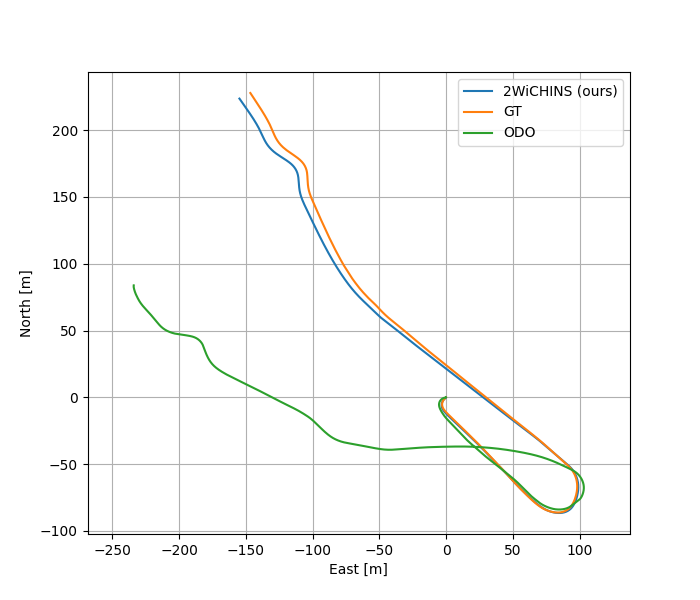}
    \includegraphics[width=0.48\linewidth]{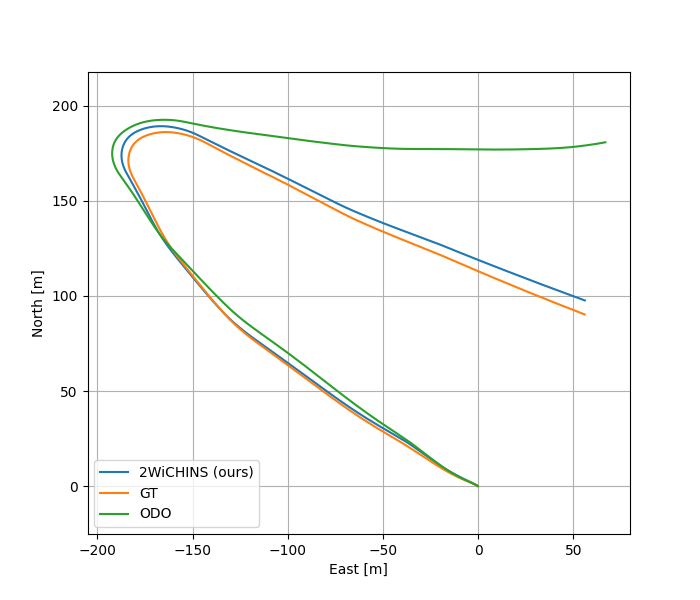}
    \caption{ The GT trajectory and the 2WiCHINS and ODO trajectory estimates.  Left:  Trajectory 1. Right: Trajectory 2.}
    \label{fig:trial1_trajectory}
\end{figure*}

\begin{figure*}
    \centering
    \includegraphics[width=0.48\linewidth]{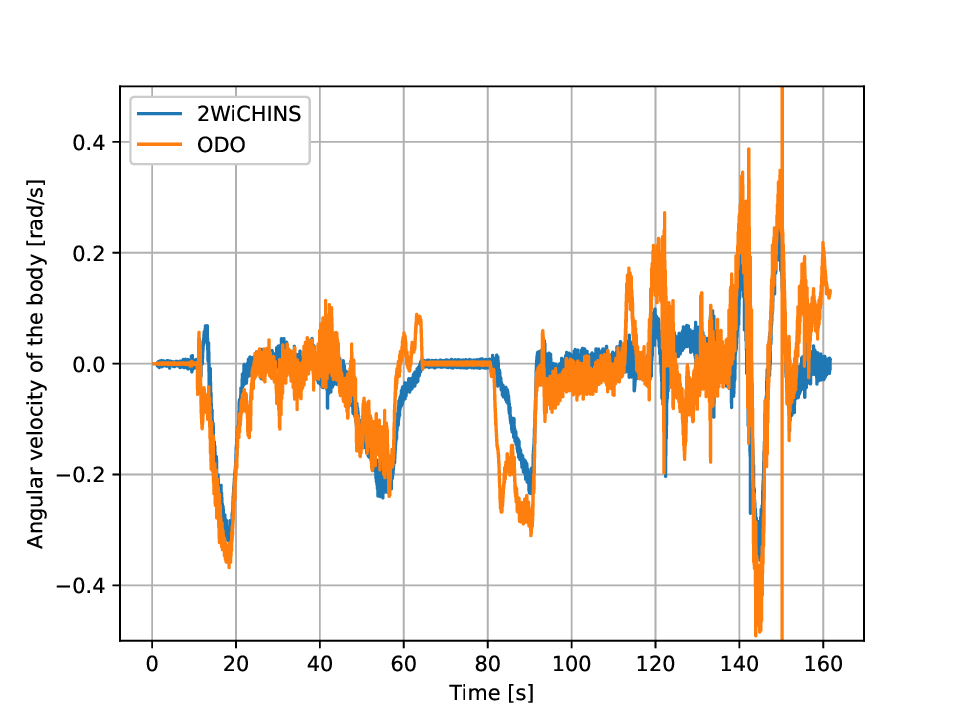}
    \includegraphics[width=0.48\linewidth]{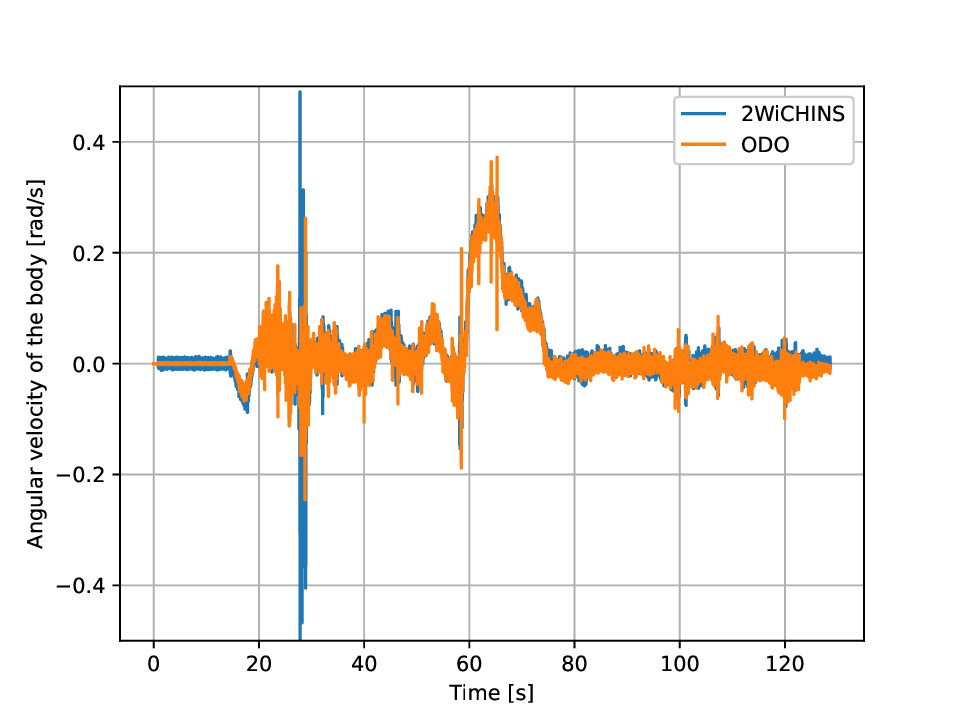}
    \caption{Estimated angular velocity around vertical axis in the body frame: comparison of our approach (2WiCHINS), with the wheel-based odometry (ODO).  Left:  Trajectory 1. Right: Trajectory 2.}
    \label{fig:trial1_gyro}
\end{figure*}

\begin{figure*}
    \centering
    \includegraphics[width=0.48\linewidth]{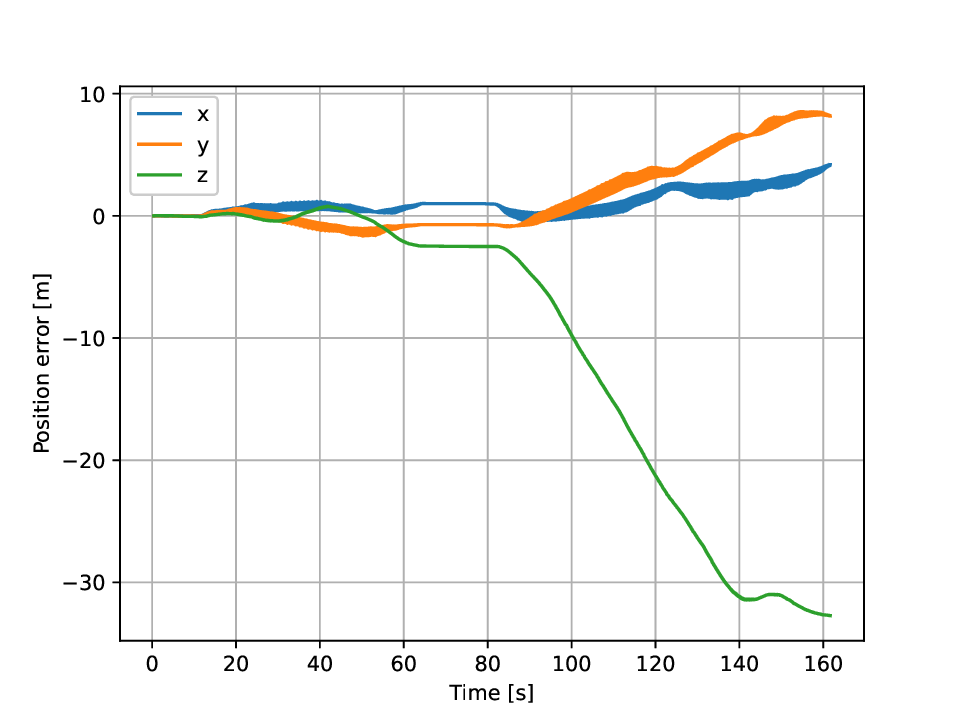}
    \includegraphics[width=0.48\linewidth]{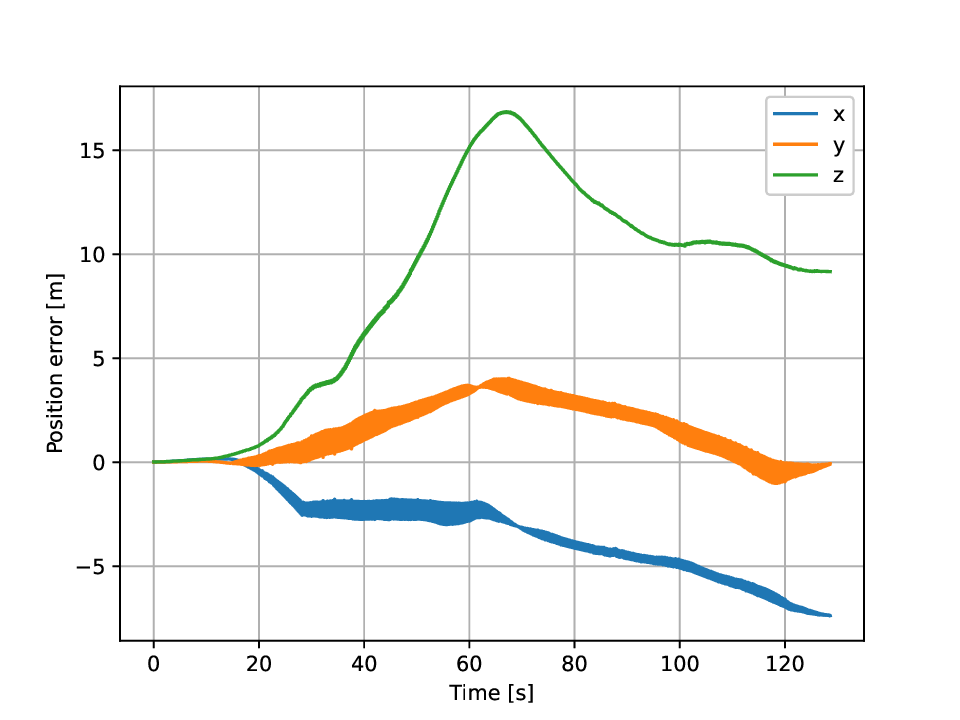}
    \caption{Position error  of the estimated position components using our approach (2WiCHINS) versus time.  Left:  Trajectory 1. Right: Trajectory 2.}
    \label{fig:trial1_pose_err}
\end{figure*}

\begin{figure*}
    \centering
    \includegraphics[width=0.48\linewidth]{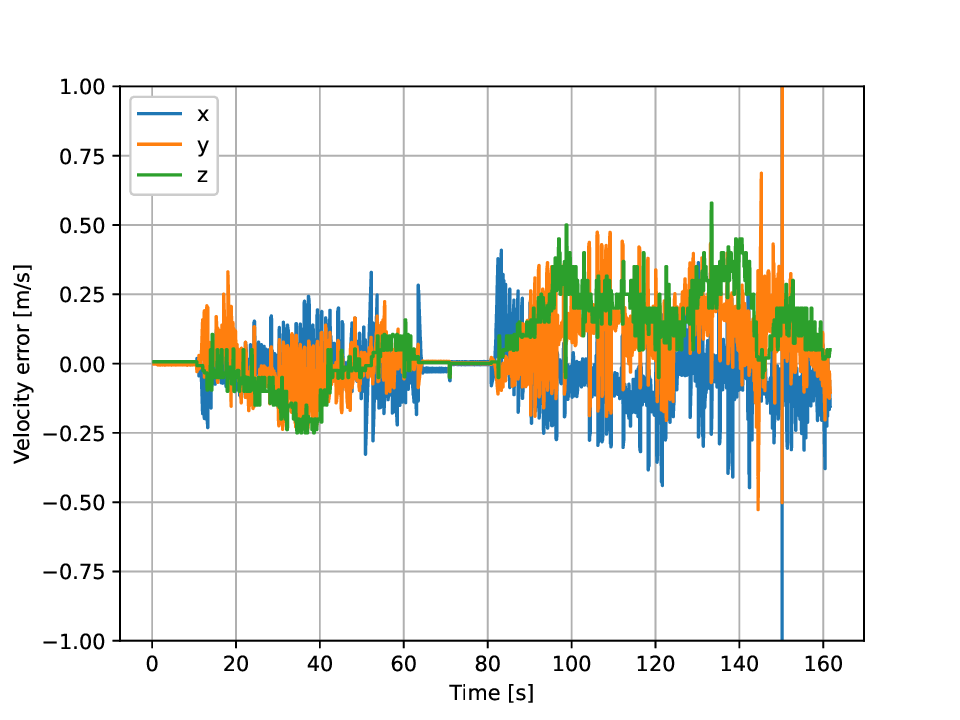}
    \includegraphics[width=0.48\linewidth]{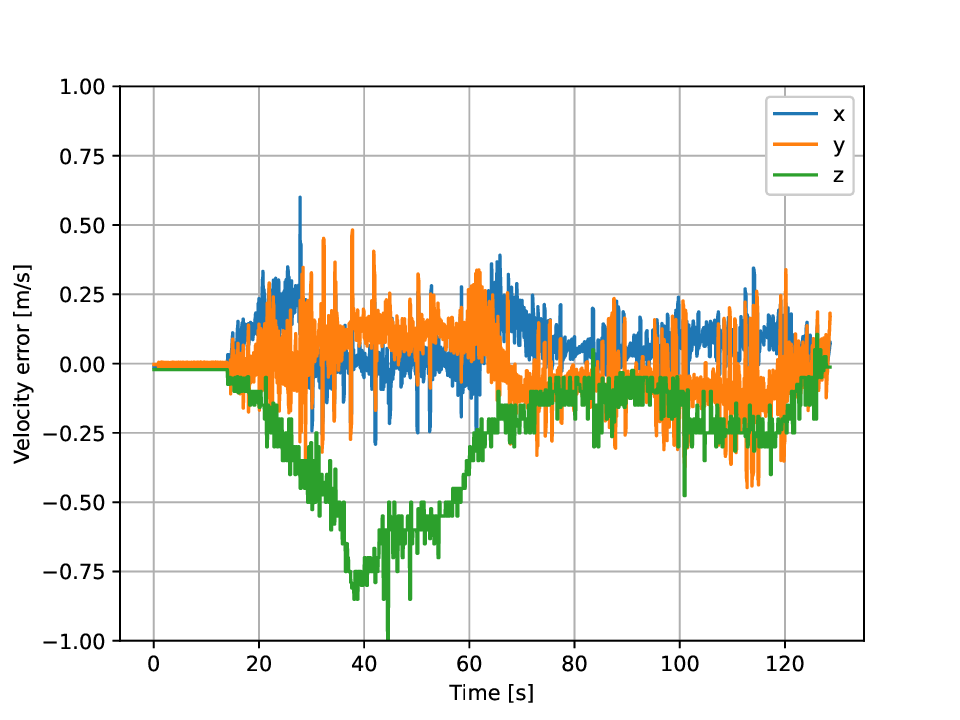}
    \caption{Velocity error  of the estimated position components using our approach (2WiCHINS) versus time. Left:  Trajectory 1. Right: Trajectory 2.}
    \label{fig:trial1_vel_err}
\end{figure*}

\noindent 
All five inertial-only approaches (described in Section~\ref{sec:Comp}) were applied on all ten trajectories (described in Section~\ref{sec:Data}).
PRMSE, VRMSE, and TDE results for all approaches and trajectories are provided in the following tables.
Table~\ref{tabcomparison} presents the PREMSE showing WiCHINS obtained dramatic improvements relative to the WMI and CMI approaches. Also, 2WiCHINS imporoved the ODO PRMSE by $87.8\%$. Finally, we observe that mounting the IMUs on all four wheels, 4WiCHINS, does not provide any significant improvement ($\approx 1\%$) compared to the 2WiCHINS setup. \\
\begin{table}[h] 
\centering
\caption{PRMSE of the estimated position vector for all five inertial approaches across 11 trajectories.}
\begin{tabular}{|c|c|c|c|c|c|}
\hline
\multicolumn{6}{|c|}{\textbf{PRMSE {[}m{]}}} \\ \hline
\textbf{Trajectory} & \textbf{2WiCHINS} & \textbf{4WiCHINS} & \textbf{ODO} & \textbf{WMI} & \textbf{CMI} \\ \hline
1  & 16.48 & 16.52 & 76.30  & 6569.99  & 3211.43  \\ \hline
2  & 10.93 & 11.00 & 41.19  & 6905.25  & 4286.54  \\ \hline
3  & 11.30 & 11.76 & 245.37 & 8677.71  & 8690.12  \\ \hline
4  & 21.51 & 21.44 & 203.32 & 3665.79  & 12592.17 \\ \hline
5  & 23.12 & 23.13 & 317.00 & 14269.49 & 30376.62 \\ \hline
6  & 3.82  & 3.84  & 5.13   & 2729.49  & 2760.46  \\ \hline
7  & 9.46  & 9.45  & 35.54  & 5342.21  & 7586.02  \\ \hline
8  & 8.28  & 8.31  & 35.05  & 6513.47  & 7290.15  \\ \hline
9  & 6.49  & 6.51  & 20.52  & 4350.39  & 4391.39  \\ \hline
10 & 3.72  & 3.88  & 5.11   & 1854.25  & 1620.97  \\ \hline
11 & 10.02 & 10.30 & 39.92  & 10517.74 & 18813.86 \\ \hline
\textbf{Average} & \textbf{11.37} & \textbf{11.46} & \textbf{93.12} & \textbf{6490.53} & \textbf{9238.16} \\ \hline
\end{tabular}
\label{tabcomparison}
\end{table}
\noindent
Similar behavior was obtained for the VRMSE as shown in Table~\ref{tab:VRMSE}. 2WiCHINS improved the ODO VRMSE by $85.2\%$. Also, for the VRMSE metric, 4WiCHINS, does not provide any significant improvement ($\approx 1\%$) compared to the 2WiCHINS setup. \\
\begin{table}[h!]
\centering
\caption{VRMSE of the estimated position vector for all five inertial approaches across 11 trajectories}
\begin{tabular}{|c|c|c|c|c|c|}
\hline
\multicolumn{6}{|c|}{\textbf{VRMSE [m/s]}} \\ \hline
\textbf{Trajectory} & \textbf{2WiCHINS} & \textbf{4WiCHINS} & \textbf{ODO} & \textbf{WMI} & \textbf{CMI} \\
\hline
1  & 0.23 & 0.22 & 1.98 & 209.17 & 208.68 \\ \hline
2  & 0.37 & 0.37 & 1.18 & 164.23 & 162.90 \\ \hline
3  & 0.96 & 0.97 & 6.70 & 201.45 & 201.22 \\ \hline
4  & 1.16 & 1.16 & 4.82 & 152.94 & 266.70 \\ \hline
5  & 1.69 & 1.69 & 5.59 & 278.87 & 419.75 \\ \hline
6  & 2.50 & 2.50 & 2.51 & 95.58  & 95.45  \\ \hline
7  & 2.03 & 2.03 & 2.24 & 114.88 & 178.18 \\ \hline
8 & 0.11 & 0.11 & 1.34 & 180.89 & 193.07 \\ \hline
9 & 0.09 & 0.09 & 1.20 & 164.04 & 164.03 \\ \hline
10 & 0.09 & 0.10 & 0.33 & 90.35  & 92.61  \\ \hline
11 & 0.17 & 0.16 & 1.01 & 350.31 & 354.51 \\ \hline
\textbf{Average} & \textbf{0.85} & \textbf{0.85} & \textbf{2.63} & \textbf{182.07} & \textbf{212.46} \\
\hline
\end{tabular}
\label{tab:VRMSE}
\end{table}
\noindent
Table~\ref{tde_table} gives the TDE metric results. Our 2WiCHINS obtained $2.31\%$ improving by a factor of $6.8\%$ the other best inertial approach (ODO).  \\
\begin{table}[h!]
\centering
\caption{TDE of the estimated trajectories for all five inertial approaches across 11 trajectories.}
\begin{tabular}{|c|c|c|c|c|c|}
\hline
\multicolumn{6}{|c|}{\textbf{TDE [\%]}} \\ \hline
\textbf{Trajectory} & \textbf{2WiCHINS} & \textbf{4WiCHINS} & \textbf{ODO} & \textbf{WMI} & \textbf{CMI} \\ \hline
1  & 3.00 & 3.01 & 13.90 & 1196.52 & 584.86 \\ \hline
2  & 2.07 & 2.08 & 7.80  & 1307.84 & 811.86 \\ \hline
3  & 2.04 & 2.12 & 44.33 & 1567.59 & 1569.83 \\ \hline
4  & 3.87 & 3.86 & 36.58 & 659.45  & 2265.23 \\ \hline
5  & 2.47 & 2.47 & 33.86 & 1524.34 & 3244.98 \\ \hline
6  & 1.84 & 1.85 & 2.47  & 1316.81 & 1331.75 \\ \hline
7  & 1.74 & 1.73 & 6.52  & 980.29  & 1392.03 \\ \hline
8  & 2.11 & 2.11 & 8.92  & 1657.71 & 1855.38 \\ \hline
9  & 2.09 & 2.09 & 6.60  & 1398.57 & 1411.75 \\ \hline
10 & 1.95 & 2.03 & 2.68  & 971.63  & 849.39  \\ \hline
11 & 2.21 & 2.27 & 8.82  & 2322.93 & 4155.19 \\ \hline
\textbf{Average} & \textbf{2.31} & \textbf{2.33} & \textbf{15.68} & \textbf{1354.88} & \textbf{1770.20} \\ \hline
\end{tabular}
\label{tde_table}
\end{table}
\noindent
To visualize the results, Trajectories 1 and 2 with our 2WiCHINS approach (best performance) and ODO (best competitive performance) trajectory estimates are shown in Figure~\ref{fig:trial1_trajectory}. The ODO approach starts to diverge after first turn, probably due to skid of the wheels, while the proposed method (2WiCHINS) allows us to compensate such errors and follow the GT trajectory. Hence, we may assume the noisy estimate of the chassis angular velocity in $z$-axis is one of the key sources of the ODO errors. 
\noindent 
This behavior is clearly visible in Figure~\ref{fig:trial1_gyro}, where the odometry-based estimate of the body angular velocity ($z$-component) is much noisier compared to WiCHINS. The trajectories produced by WMI and CMI approaches are not shown in the figure, since their errors are by the order of magnitude higher, as they start to diverge almost instantly. Such a behavior is caused mainly by the high bias instability and noise of the inertial sensors (see Table \ref{tab:gyro_accel}). \\
\noindent The error of the estimated position and velocity components obtained by our method (2WiCHINS) are shown in Figure~\ref{fig:trial1_pose_err} and Figure~\ref{fig:trial1_vel_err}, respectively. The main portion of the error in in the vertical $z$-axis. For wheeled robots in most applications only the horizontal position components are relevant, hence the vertical component can be ignored. The velocity errors are effectively compensated to the level when they do not show almost any predictable patterns. Again, the significant part of the error is in the $z$-axis, which can be ignored in the most applications.   

\section{Conclusion}\label{sec:con} 
\noindent Autonomous vehicles and wheeled robots operating in environments with degraded or denied GNSS signals require robust and accurate navigation based only on inertial sensors. Yet, the inherent drift of the inertial solution presents a critical challenge in such scenarios.
To fill this gap, we presented WiCHINS, a novel approach that synergistically combines the complementary strengths of wheel-mounted and chassis-mounted IMUs. Our core contribution is a three-stage framework specifically designed to leverage the distinct motion characteristics observed at each sensor location for enhanced estimation. The wheel-mounted IMUs provide strong translational information and the chassis-mounted IMU offers stability and comprehensive orientation information. \\
\noindent
The performance of our proposed WiCHINS method was rigorously evaluated using a comprehensive dataset featuring five IMUs and a total recording time of 228.6 minutes and against other inertial-based approaches. Our experimental results conclusively demonstrate the effectiveness of the fusion strategy: utilizing just two-wheel IMUs and one chassis IMU. WiCHINS achieved an average 3D position error of 11.37 meters which is $2.4\%$ of the average travelled distance. This result confirms the superiority of our combined sensor configuration and multi-stage filtering architecture, enabling robust navigation in challenging environments. \\
\noindent
While the WiCHINS framework significantly reduces position drift, the pure inertial nature of the solution introduces several limitations. The core OriEKF for body orientation estimation relies on a body-mounted IMU and, in our experimental setup, did not utilize magnetometer readings. Without an external absolute heading reference (magnetometer), the yaw (heading) angle is prone to accumulating drift over long periods, which directly impacts the overall navigation accuracy. In addition, the PosEKF update step is derived from inverse kinematics which assumes no-skid conditions between the wheel and the ground. The performance is expected to degrade in scenarios involving aggressive maneuvers, severe wheel slip, or operation on slippery terrains (e.g., ice, mud).
The primary strength of the WiCHINS framework is its ability to successfully mitigate the rapid drift characteristic of pure inertial solutions and offer superior accuracy. The three-stage EKF efficiently exploits the complementary kinematic information, the rotational stability from the chassis IMU and the high-frequency translational constraints from the wheel IMUs, leading to a highly optimized navigation solution. As a purely inertial system, WiCHINS offers an inherently robust and self-contained backup or primary navigation source that is immune to signal blockage in environments where GNSS, visual, or radio signals are unavailable.
By bridging the pure-inertial performance gap, this research provides a reliable, self-contained, and high-integrity alternative to conventional inertial-based solutions, thus increasing the safety, reliability, and operational envelope of autonomous vehicles and mobile robots.
\section*{Acknowledgment}
\noindent
The authors are grateful to the support of the project APVV SK-IL-RD-23-0002: Advanced Localization Sensors and Techniques for Autonomous Vehicles and Robots supported by Slovak Research and Development Agency, and to the support of The Israeli Ministry of Science, Technology and Space under grant number 06725.
\ifCLASSOPTIONcaptionsoff
  \newpage
\fi



%
\newpage
\bibliographystyle{ieeetr}
\bibliography{bio}

\end{document}